\documentclass[runningheads]{llncs}

\usepackage{eccv}

\usepackage{eccvabbrv}

\usepackage{graphicx}
\usepackage{booktabs}

\usepackage[accsupp]{axessibility}  %

\usepackage[pagebackref,breaklinks,colorlinks,citecolor=eccvblue]{hyperref}
\usepackage{graphicx}
\usepackage{amsmath}
\usepackage{amssymb}
\usepackage{booktabs}
\usepackage{multirow}
\usepackage{tabularx}
\usepackage{float}
\usepackage{afterpage}
\usepackage{wrapfig}
\usepackage{algorithm}
\usepackage{algorithmic}
\usepackage{tikz}
\usepackage{pifont}
\usepackage{wrapfig,lipsum,booktabs}
\usepackage{orcidlink}

\begin{document}

\title{Unsupervised Multimodal Deepfake Detection Using Intra- and Cross-Modal Inconsistencies} 

\titlerunning{Deepfake Detection Using Intra- and Cross-Modal Inconsistencies}

\author{Mulin Tian\inst{1,2}$^*$\orcidlink{0000-1111-2222-3333} \and
Mahyar Khayatkhoei\inst{1,}$^*$\orcidlink{1111-2222-3333-4444} \and
Joe Mathai\inst{1,}$^*$\orcidlink{2222--3333-4444-5555} \and \\ 
Wael AbdAlmageed\inst{3}\orcidlink{3333-4444-5555--6666}}

\authorrunning{Tian et al.}

\institute{USC Information Sciences Institute, Marina del Rey, USA \and
USC Ming Hsieh Department of Electrical and Computer Engineering, Los Angeles, USA \and
Holcombe Department of Electrical and Computer Engineering, Clemson University, Clemson, USA
}

\maketitle
\def\thefootnote{*}\footnotetext{These authors contributed equally to this work.}
\def\thefootnote{\arabic{footnote}}

\begin{abstract}
Deepfake videos present an increasing threat to society with potentially negative impact on criminal justice, democracy, and personal safety and privacy. Meanwhile, detecting deepfakes, at scale, remains a very challenging task that often requires labeled training data from existing deepfake generation methods. Further, even the most accurate supervised deepfake detection methods do not generalize to deepfakes generated using new generation methods. In this paper, we propose a novel unsupervised method for detecting deepfake videos by directly identifying intra-modal and cross-modal inconsistency between video segments. The fundamental hypothesis behind the proposed detection method is that motion or identity inconsistencies are inevitable in deepfake videos. We will mathematically and empirically support this hypothesis, and then proceed to constructing our method grounded in our theoretical analysis. Our proposed method outperforms prior state-of-the-art unsupervised deepfake detection methods on the challenging FakeAVCeleb dataset, and also has several additional advantages: it is scalable because it does not require pristine (real) samples for each identity during inference and therefore can apply to arbitrarily many identities, generalizable because it is trained only on real videos and therefore does not rely on a particular deepfake method, reliable because it does not rely on any likelihood estimation in high dimensions, and explainable because it can pinpoint the exact location of modality inconsistencies which are then verifiable by a human expert.


\end{abstract}

\begin{figure*}[!htb]
  \begin{center}
    \includegraphics[trim=0 0 0 0, clip, width=0.6\textwidth]{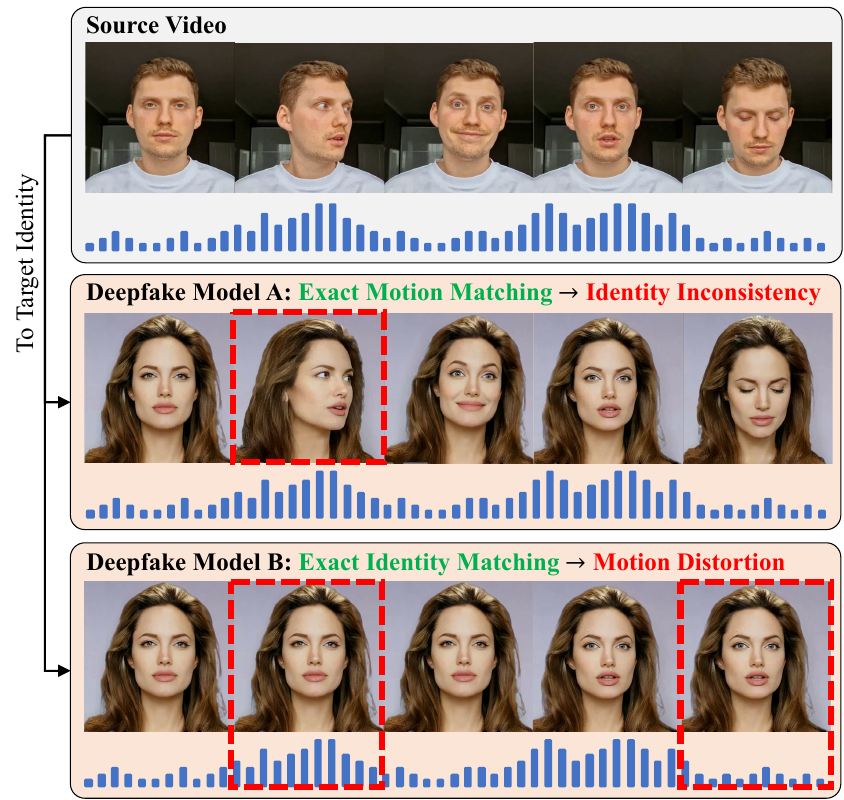}
    \caption{When transferring the motion of the source video \emph{(top)} to the target identity \emph{(Angelina Jolie)}, the deepfake generation method~\cite{drobyshev2022megaportraits} faces a trade-off: \emph{(middle)} matching motion exactly results in some frames having the wrong identity which can be detected by looking for intra-modal identity inconsistency; or \emph{(bottom)} matching identity exactly results in motion distortion which can be detected by looking for video cross-modal inconsistency with audio (\eg, the lips do not move at moments where audio magnitude shows speaking, and vice versa). Red boxes show inconsistencies.}
    \label{fig:motivation}
  \end{center}
  \vspace{-.3in}
\end{figure*}  

\section{Introduction}
\label{sec:intro}
The rapid advancement of generative deep learning~\cite{rombach2022stable-diffusion, ramesh2022dalle2, karras2021stylegan3}, fueled by faster and cheaper compute power and exploding data availability, has blurred the line between fact and fiction. In particular, \textit{deepfakes}\footnote{We use \emph{deepfake} as an umbrella term to refer to fake video generation in general including face-swapping, face reenactment, face generation, etc.}---videos in which the motion of a source video is transferred to a target identity such that it appears to say the words uttered by the source---are becoming increasingly hard to distinguish from real videos~\cite{drobyshev2022megaportraits}. Since existing deepfake detection methods themselves can be used as part of the objective in the generation process to improve the deepfake quality, this rapid advancement in generative deep learning leads to a daunting question: \emph{will machines eventually be able to imitate any person without leaving any trace?} Despite the alarming empirical evidence~\cite{carlini2020evading, huang2020fakepolisher}, we conjecture that the answer could be negative; that is, deepfakes might always contain a \emph{detectable inevitable trace}, as illustrated in \cref{fig:motivation}, which explains the the observation that has led to our conjecture. While the synthesized video (middle row) appears very realistic in each frame and its motion almost exactly matches those of the source, a closer look reveals that at certain frames the identity of the person has noticeably changed (depicted by a red dashed box).
We conjecture that this phenomena, which occurs in various deepfake videos~\cite{fakeav}, is a consequence of a fundamental property of facial motion and identity, that \emph{facial motion and identity are not independent variables}. Consequently, a deepfake model must either not perfectly transfer the motion  (leading to motion inconsistencies in the target video) or perfectly transfer the motion and with it partially transfer the identity as well (leading to identity inconsistencies in the target video). The goal of this paper is to mathematically and empirically validate this conjecture (\cref{sec:theory}), and to develop a new method for explicitly detecting the conjectured inconsistencies (\cref{sec:methods}).

Our proposed \emph{unsupervised} deepfake detection method is designed to simultaneously detect intra- and cross-modal inconsistencies within a given video. Note that while extracting consistency-based \emph{features} have been explored in prior deepfake detection (\ie, learning joint features from video-audio based on a contrastive loss~\cite{Audio_visual_POI_deepfake_detection, Audio-Visual_Dissonance-based_Deepfake_Detection_and_Localization}), our work is the first to directly identify and pinpoint inconsistencies within a given video, \emph{without relying on comparison to any other pristine (real) video}. This distinction makes our method explainable (since the pinpointed inconsistencies of a video can be directly provided to a human expert as explanation of fakeness), and scalable to many identities (since it does not need pristine videos of identities to compare against at inference time). Our method outperforms prior state-of-the-art unsupervised detection methods on the challenging FakeAVCeleb~\cite{fakeav} dataset (achieving a new best average AUC of $96.81\%$). We summarize the novelty and  practical advantages of our method compared to existing deepfake detection methods as follows.

\begin{enumerate}
    \item \textbf{Inevitable inconsistencies:} we provide mathematical and empirical evidence for that deepfake generation methods will inevitably leave generation artifacts/traces due to a trade-off between identity and motion. To the best of our knowledge, this argument does not exist in the literature.
    
    \item \textbf{Generalizability:} compared to supervised methods~\cite{Noiseprint, artifacts, artifacts2,masi2020two} which rely on existing deepfake generation models, our proposed method is unsupervised and is trained solely on real videos, thereby does not rely on the particular of artifacts of existing deepfake generation models.

    \item \textbf{Scalability:} compared to POI-Forensics~\cite{Audio_visual_POI_deepfake_detection}, which requires access to a set of pristine reference samples for each identity at inference time in order to evaluate the given video, our method does not require any pristine samples, thereby can scale to arbitrarily many new identities.

    \item \textbf{Reliability:} compared to AV-Anomaly~\cite{self_supervised_video_forensics_by_audiovisual_anomaly_detection} which relies on generative modeling to estimate the likelihood of real data (making it susceptible to the known unreliability of likelihood estimation in high-dimensions~\cite{nalisnick2018gms-dont-know}), our method does not use any likelihood estimation, and instead directly compares a given video with itself to find mismatching regions.

    \item \textbf{Explainability:} our method is explainable by design, since to detect whether a given video is fake, it must discover that a portion of the video is inconsistent with another portion of the same video. These two inconsistent portions can be provided to a human expert for verification of inconsistency.
\end{enumerate}

\section{Related Work}
\label{sec:related_works}
\noindent\textbf{Deepfake Generation.}
Along with the development of generative models, there are many ways to generate fake videos for both the visual and auditory parts of a video. Suwajanakorn \etal. \cite{SynthesizingObama} proposed an approach to produce videos of President Obama with realistic lip movement from a given audio. Utilizing face-swapping \cite{fakeav} and lip-syncing, GANimation\cite{GANimation}, FSGAN\cite{fsgan}, Wav2lip\cite{wav2lip}, Face2Face\cite{face2face} can generate deep fake videos with better quality. SV2TTS\cite{SV2TTS} can even generate an audio speech for a different person by a given piece of text. To deal with the ethical and security issues brought by deepfake techniques, deepfake detection methods are proposed.

\noindent\textbf{Unimodal Deepfake Detection.} 
The unimodal deepfake detection methods focus on detecting the artifacts in visual part of a video. Li \etal. \cite{exposing_deepfake_by_face_warping_artifacts} claimed that the artifacts are left in deepfake videos, where such artifacts can be effectively captured by a convolution neural network (CNN). Built upon \cite{exposing_deepfake_by_face_warping_artifacts}, Güera \etal. \cite{deepfake_video_detection_using_RNNs} incorporated a recurrent neural network (RNN) to detect Deepfake frames by feeding the RNN with the features extracted from each frame by CNN. Except raw videos, Yang \etal. \cite{exposing_deepfake_using_head_positions} suggested that the fake videos can be detected by analyzing the estimated head position corresponding to the facial landmarks. 

\noindent\textbf{Multimodal Deepfake Detection.}
Several approaches have recently focused on multimodal deepfake detection by utilizing both visual and auditory information. The key of multimodal method is to find the way to measure the dissimilarity between audio and video features \cite{an_audiovideo_deepfake_detection_method_using_affective_cues}. Li \etal.\cite{an_audiovideo_deepfake_detection_method_using_affective_cues} proposed detecting Deepfake videos by analyzing audio-visual cues and perceived emotions which was extracted from Memory Fusion Network (MFN)\cite{MFN}.  Chugn \etal. \cite{Audio-Visual_Dissonance-based_Deepfake_Detection_and_Localization} incorporated the contrastive learning as the objective function and Modality Dissonance Score (MDS) to measure audio-video dissimilarity. Similarly, Hshmi \etal. \cite{Multimodal_Forgery_Detection_Using_Ensemble_Learning} fed both visual features extracted from ResNet\cite{resnet} and audio Mel-spectral features \cite{mel_sprc} into a fusion network to produce the final prediction score. In Zhou. \cite{joint_audio_visual_deepfake_detection}, similar idea was implemented by checking if the concurrency property of audio and visual features is broken. Shahzad~\etal~\cite{lip_sync_matters} proposed a multimodal approach using Wav2Lip \cite{wav2lip} to generate a corresponding audio semantic features which help the model to distinguish between original and Deepfake videos. Aforementioned methods rely on training the model with groundtruth labels of fake and real videos. In contrast, AV-Anomaly \cite{self_supervised_video_forensics_by_audiovisual_anomaly_detection}, first learns a joint audio-visual segment feature in a contrastive setting from real videos, and then trains a likelihood estimator on those features extracted from real videos (an autoregressive generative model of sequences of synchronization characteristics). The likelihood estimator is then used to score a query video's likelihood of being real. Another unsupervised method, POI-Forensics \cite{Audio_visual_POI_deepfake_detection}, proposed an audio-visual identification verification approach, where a model is trained to extract features that indicate the identities of videos through contrastive learning on real videos, and then during inference, these features are used to compare a query video to a set of existing pristine (real) videos, to determine a fakeness score. In contrast to these methods which use audio-visual consistency as a mean to learn representative features for later processes (likelihood estimation or comparison to pristine videos), we aim to theoretically motivate and then develop a method that directly identifies inconsistencies in a query video.

\section{Theoretical Motivation}
\label{sec:theory}
In this section, we provide the theoretical justification for our conjecture that changing the motion of a target video to match the motion of the source video will inevitably introduce  intra- or cross-modal inconsistencies, as illustrated in~\cref{fig:motivation}. To formalize the identity, the facial motion and the specific video, let $Y$ be a random variable representing the identity of a person (with a sample space of all human identities), $M$ a random variable representing facial motion (with a sample space of all realistic facial motions within a specific time-window), and $V$ a random variable representing a video of a person (with a sample space $\mathcal{V}$ of all talking-head videos of a specific duration). We model the deepfake generation process as a mapping $g: \mathcal{V} \rightarrow \mathcal{V}$ which induces a random variable $V_g$, and consider the following notions (where $I$ is mutual information and $H$ entropy):
\begin{itemize}
    \item $I(V_g;Y)$ measures the correspondence between a video and an identity, where a larger mutual information between video and identity indicates that the video has a distinct and consistent identity.
    \item $I(V_g;M)$ measures the correspondence between video and the facial motion, where a smaller mutual information between video and motion indicates that the motion can be transferred across videos.
    \item $H(Y|M)$ measures identity dependence on the motion, where a smaller conditional entropy (\ie uncertainty) indicates that the motion is more predictive of the identity.
\end{itemize}
In this theoretical framework, we can state the objective of the deepfake generation method $g$ as achieving a high $I(V_g;Y)$ (\ie, the target video can have the target identity) while simultaneously achieving a low $I(V_g;M)$ (\ie, the target video can have the desired motion of the source video without distortion). The relationship between these two objectives (maximizing $I(V_g;Y)$ and minimizing $I(V_g;M)$) can be explained by the following inequality which holds for any three random variables~\cite{li2023bias-bounds}: 
\begin{align}
\label{th:bound}
    0 \leq I(V_g;Y) \leq I(V_g;M) + H(Y|M).
\end{align}
In the above inequality, we observe that if the motion is predictive of the identity ($H(Y|M)\rightarrow 0$), then the fake generation process faces a trade-off: if it learns to precisely transfer the motion among videos ($I(V_g;M) \rightarrow 0$), then it will inevitably break the identity consistency in the generated videos ($I(V_g;Y) \rightarrow 0$).

\textbf{Is motion predictive of identity in practice?} In~\cref{th:bound}, we observed that \emph{if the motion is predictive of identity in real videos} ($H(Y|M)\rightarrow 0$), then any deepfake generation method will be in a trade-off between motion consistency and identity consistency. However, is motion predictive of identity in real videos? Recent works~\cite{agarwal2020detect-deepfake-behavior, Audio_visual_POI_deepfake_detection} utilize temporal features for identity recognition and verification, indicating that the motion is indeed predictive of identity. However, to our knowledge, the existing evidence does not explicitly separate motion from pose and appearance features, and therefore we conduct an experiment to directly observe how predictive the motion is of the identity. More concretely, we model motion in a video by extracting the difference between 3D facial landmarks for all pairs of consecutive frames in the video, and concatenate these differences into a long input feature vector, denoted the \textit{motion vector} for a video. Then, we investigate the accuracy of an identity classifier trained on motion vectors. Note that, unlike methods that extract temporal features, the simple \textit{motion vector} is intentionally crafted such that it does not include any pose or appearance features, that could have been unintentionally learned as part of the temporal features. Since the motion vectors do not contain any appearance and pose information, the accuracy of this classifier serves as an indication of whether motion is predictive of identity. The experiment is conducted on VoxCeleb2~\cite{VoxCeleb2} dataset, where we partition the videos in the official training set\footnote{The official testing set does not contain as many identities and lacks video labels.}, containing 5994 unique identities, into a training and validation set (a $80\%$-$20\%$ partition), making sure there is no overlap between the videos in training and validation sets. We train a CNN~\cite{resnet} backbone on the extracted motion vectors. This model achieves a validation accuracy of $9.94\%$ which is a 596-fold improvement over random guess accuracy. We expect this accuracy can be further improved by a more complex model, but the observed accuracy suffices for our current goal of providing evidence for that motion is predictive of identity in real videos.

Given that the motion is predictive of identity, and that the inequality in~\cref{th:bound} is invariant to the choice of model $g$, we argue that fake generation processes, in general, must either sacrifice identity consistency within the generated video, or the exact transfer of the motion from the source video to the target video. This in turn means that any fake generation process will inevitably leave artifacts as a result of the trade-off between identity consistency and the accuracy of motion transfer. To further support our theoretical claim, we also conduct an identity prediction from motion experiment on the FakeAVCeleb~\cite{fakeav} (213 identities). In~\cref{tab:motion_fakeav}, we observe that in real videos (trained and tested on reals) identity is much more predictable from motion compared to in deepfake videos (trained and tested on deepfakes), showing that deepfake generation breaks the motion-identity interdependence, consistent with our theoretical claim that motion or identity inconsistencies are inevitable in deepfakes.

{
\setlength{\tabcolsep}{8pt}
\begin{table*}[t]
  \caption{Accuracy(\%) of 213-class identity classification from motion on FakeAVCeleb.}
   \centering
   \begin{tabular}{cccc}
\toprule
 & & \multicolumn{2}{c}{Training Set} \\
& & Real & Fake \\
\midrule
\multicolumn{1}{c}{\multirow{2}{*}{Testing Set}}           & Real & 14.56 & 0.65 \\ 
& Fake & 0.90 & 2.62 \\
\bottomrule
\end{tabular}

   \label{tab:motion_fakeav}
 \end{table*}
 }

\begin{figure*}[t]
    \centering
    \includegraphics[width=1\textwidth]{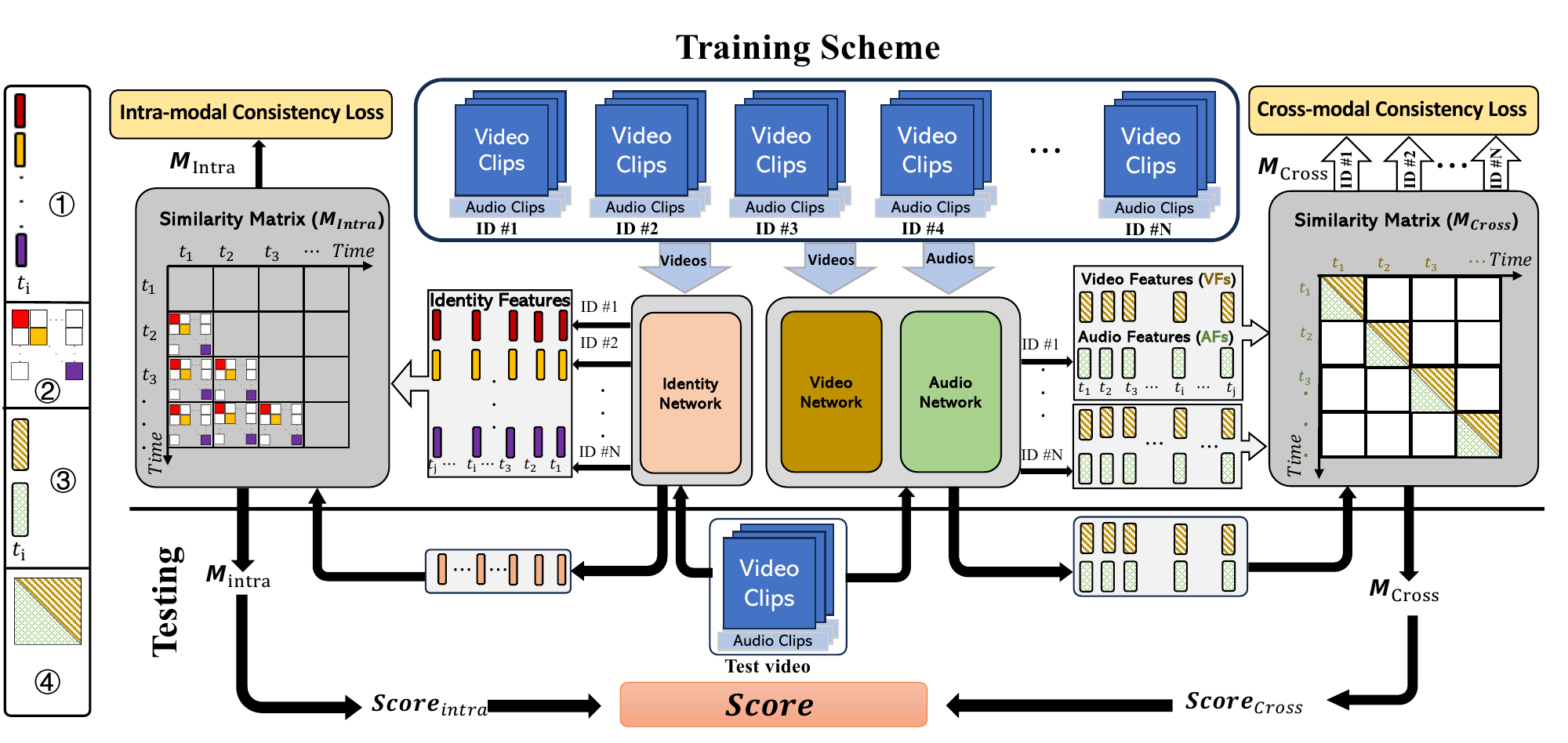}
    \caption{
    Training and testing scheme for intra-modal consistency and cross-modal consistency methods. For each training batch, we take multiple fixed-size video and audio clips of $N$ distinct identities and feed them into our networks. {\fontsize{12}{14}\selectfont \ding{172}}{\fontsize{10}{10} is an output (feature vector) of all identities extracted from the identity network at time-window $t_a$, The similarity matrix computed on time dimension is given in the gray box on the left,} each element {\fontsize{12}{14}\selectfont \ding{173}} represents the similarity matrix of {\fontsize{12}{14}\selectfont \ding{172}} on a specific time-window pair {\fontsize{10}{10}$(t_{a},t_{b})$}. A intra-modal consistency loss for identity network training is calculated based on this tensor. {\fontsize{12}{14}\selectfont \ding{174}} denotes the feature vector generated by video and audio network at time-window {\fontsize{10}{10}$t_{a}$}. The features of each individual across multiple time windows are used to generate their corresponding similarity matrix. A cross-modal consistency loss for video and audio network training is calculated from these N 2-dimensional matrices.
    }
    \label{fig:model}
\end{figure*}

\section{Deepfake Detection}
\label{sec:methods}
\newcommand{\crosscircle}{
    \begin{tikzpicture}[scale=0.1]
        \draw (0,0) circle [radius=1];
        \draw (-0.7,-0.7) -- (0.7,0.7);
        \draw (-0.7,0.7) -- (0.7,-0.7);
    \end{tikzpicture}
}
\newcommand{\dotcircle}{
    \begin{tikzpicture}[scale=0.1]
        \draw (0,0) circle [radius=1];
        \fill (0,0) circle [radius=0.3];
    \end{tikzpicture}
}

Motivated by the theoretical analysis in~\cref{sec:theory}, we propose a deepfake detection method based on the premise of inevitable inconsistencies in terms of ---(1) identity inconsistency in the generated video (due to small $I(V_g;Y)$) by comparing different video segment pairs in the video, and (2)  motion distortions in the generated video (due to large $I(V_g;M)$) by comparing video and audio segments.

\noindent\textbf{Notation and Training Setup.} Assume a training batch that includes multiple videos $\{v_i\}_{i=1}^N$ for $N$ distinct \emph{real} identities, each video having $T$ different time-window (a time-window is a fixed duration of time). Each video goes through three mappings parameterized by neural networks to extract identity, visual and audio features. Let $\mu: \mathcal{V}\times\{1\dots T\} \rightarrow \{y \in \mathbb{R}^d: ||y||_2 = 1\}$ define a mapping between a video and its corresponding identity features at a time-window $t\in [1,T]$, similarly $\gamma: \mathcal{V}\times\{1\dots T\} \rightarrow \{y \in \mathbb{R}^d: ||y||_2 = 1\}$ define a mapping between a video and its corresponding visual features at a time-window, and $\alpha: \mathcal{V}\times\{1\dots T\} \rightarrow \{y \in \mathbb{R}^d: ||y||_2 = 1\}$ define a mapping between a video and its corresponding audio features at a time-window. Whenever referring to a feature extracted from the video of identity $i$ at time-window $t$, we summarize $\mu(v_i, t)$ into $\mu_i(t)$ for brevity. The dot product is defined as $\langle\cdot,\cdot\rangle$.

\noindent\textbf{Intra-modal Consistency Loss.}
In order to detect identity inconsistencies, we must learn an identity feature extractor $\mu$ that is sensitive to slight changes of identity within any video (due to a small $I(V_g;Y)$ as described in~\cref{sec:theory}). Therefore, we propose the intra-modal consistency loss, whose objective is to guide $\mu$ to learn maximally divergent feature vectors between different identities, and maximally convergent feature vectors between different observations/samples of the same identity. We use $\mu$ to extract identify features for each sample in the batch as illustrated in the left side of~\cref{fig:model}. We then measure the similarity between all pairs of identity vectors $\langle\mu_i(t),\mu_j(q)\rangle$, where $i,j$ are identity indices and $t,q$ time-window indices, resulting in a $T \times T \times N \times N$ similarity tensor. 
\cref{eq:self-consistency-loss} shows the overall \emph{intra-modal consistency loss} $L\,_{\dotcircle}$:
{
\begin{align}\label{eq:self-consistency-loss}
L\,_{\dotcircle} = \frac{-1}{I \times T^2}\sum_{t=1}^{T} \sum_{q=1}^{T} \sum_{i=1}^{N} \log\left(\frac{\exp(\langle\mu_{i}(t),\mu_{i}(q)\rangle/\tau)}{\sum_{j=1}^{N}\exp(\langle\mu_{i}(t),\mu_{j}(q)\rangle/\tau)}\right)
\end{align}
}
where $\tau$ is the temperature used to control the scale of similarity measurements. The left part of~\cref{fig:model}  illustrates how we find the dissimilarity with given features from different identities (small white squares in {\fontsize{12}{14}\selectfont \ding{173}}-\cref{fig:model}) and the similarity with given features from the same identity (small colored squares in {\fontsize{12}{14}\selectfont \ding{173}}-\cref{fig:model}). The objective is to make the colored dot-products increase and white dot-products decrease such that vectors of similar identities becoming closer to each other than to vectors of dissimilar identities, across all pairs of time-windows.

At inference time, the  video is divided into $T$ fixed sized windows. As illustrated by the black arrows in the left part of the testing flow in~\cref{fig:model}, we  use the trained $\mu$ to compute the $T\times T$ similarity matrix of identity features. The intra-modal consistency score of the test video using intra-modal consistency method ({
\normalsize 
\begin{math}
\begin{aligned}
Score\,_{\dotcircle} 
\end{aligned}
\end{math}
}) is the $n^{th}$ percentile of this similarity matrix.

\noindent\textbf{Cross-modal Consistency Loss.}
In order to detect motion inconsistencies, we attempt to learn a visual feature extractor $\gamma$ and an audio feature extractor $\alpha$ that are sensitive to slight mismatches between audio and video of a speaking person (as a surrogate to motion inconsistency due to a large $I(V_g;Y)$ as described in~\cref{sec:theory}). Therefore, we propose the cross-modal consistency loss, whose objective is to guide $\gamma$ and $\alpha$ to learn maximally divergent feature vectors between video and audio at different time windows, and maximally convergent feature vectors between video and audio at the same time windows. We use $\gamma$ and $\alpha$ to extract visual and audio features for each sample in the batch as illustrated in the right side of~\cref{fig:model}. We then measure the similarity between all pairs of audio-video feature vectors $\langle\gamma_{i}(t),\alpha_{i}(q)\rangle$, where $i$ is the identity index and $t, q$ are time-window indices, resulting in a $N \times T \times T$ similarity tensor. \cref{eq:cross-consistency-loss} shows the overall \emph{cross-modal consistency loss} $L\,_{\crosscircle}$:

\begin{align}\label{eq:cross-consistency-loss}
\begin{split}
    L\,_{\crosscircle} = \frac{-1}{N\times T}\sum_{i=1}^{N} \sum_{t=1}^{T} \left(\right. 
    &\log\left( \frac{\exp( \langle\gamma_{i}(t), \alpha_{i}(t)\rangle/\tau' )}{\sum_{q=1}^{T} \exp( \langle\gamma_{i}(t),\alpha_{i}(q)\rangle/\tau' )} \right)\\
    + &\log\left( \frac{\exp( \langle\gamma_{i}(t), \alpha_{i}(t)\rangle/\tau' )}{\sum_{q=1}^{T} \exp( \langle\gamma_{i}(q),\alpha_{i}(t)\rangle /\tau' )} \right) \left.\right)\\
\end{split}
\end{align}
where $\tau'$ is the temperature used to control the scale of similarity measurements. The right part of~\cref{fig:model} intuitively illustrates how we find the dissimilarity with given audio and video features at different time-windows (white squares in the gray block-\cref{fig:model}) and the similarity with given audio and video features from the same time-window (squares with green and brown filling as marked as {\fontsize{12}{14}\selectfont \ding{175}}-\cref{fig:model}). The objective is to make the colored dot-products increase and white dot-products decrease such that vectors of corresponding audio-visual time-windows become closer to each other than to all other audio or video vectors of different time-window for each identity.

At inference/testing time, the testing video is divided into $T$ fixed sized windows, and following the black arrows in the right part of the testing flow in~\cref{fig:model}, we will use the trained $\gamma$ and $\alpha$ to compute the $T\times T$ similarity matrix of visual-audio features. The score of the testing video ({
\normalsize 
\begin{math}
\begin{aligned}
Score\,_{\crosscircle} 
\end{aligned}
\end{math}
}) is determined by calculating the average value of its diagonal elements. We take the sum of these two scores as the final deepfake detection score:
{
\normalsize 
\begin{equation}\label{eq:score}
\begin{aligned}
Score = Score\,_{\crosscircle} + Score\,_{\dotcircle} 
\end{aligned}
\end{equation}
}

\section{Experimental Evaluations}
\label{sec:results}

\textbf{Datasets.} We use three data sets for training and evaluating our method---(1) VoxCeleb2, (2) FakeAVCeleb and (3) KoDF. VoxCeleb2~\cite{VoxCeleb2} is used for training, since the training process of our model requires a substantial amount of speaking videos from various individuals, and VoxCeleb2  meets the requirements. VoxCeleb2 includes over 6000 different identities, each with videos of that identity speaking in multiple scenarios. Leveraging this characteristic, it is feasible to construct a substantial number of video and audio windows with fixed size.

FakeAVCeleb \cite{fakeav} is used to evaluate the performance of our model, which consists of 500 identities taken from VoxCeleb2, where both the visual and audio content of the videos of these identities is tampered. We excluded all videos of the 500 identities present in FakeAVCeleb (used for evaluation) from our training dataset (VoxCeleb2) in order to prevent our method from learning features specific to these identities. FakeAVCeleb dataset allows us to thoroughly test our model's ability to discern between different modalities of forgery. Specifically, FakeAVCeleb utilizes Wav2Lip \cite{wav2lip}, faceswap \cite{fakeav} and fsgan \cite{fsgan} for video manipulation, along with SV2TTS \cite{SV2TTS} for audio manipulation. The manipulated videos are categorized into four groups: RealVideo-RealAudio, RealVideo-FakeAudio, FakeVideo-RealAudio, and FakeVideo-FakeAudio. The first two categories contain 500 videos each, and the others have about 10000 videos each.

KoDF~\cite{KoDF} is a deepfake dataset of Korean speaking videos, which we utilize for evaluating the generalization of our method to people speaking in other languages (which indicates out-of-domain generalization to audio features). The manipulation techniques employed in KoDF include face swapping and face reenactment methods. Additionally, there is a dedicated section within the dataset where both real and fake videos have been subjected to adversarial attacks. We subsequently used the data from this section to evaluate the robustness of our model under adversarial attacks. For those real and fake videos in the KoDF dataset that had not been manipulated with adversarial attacks, we attempted to download them, but encountered persistent issues during the decoding process. Consequently, the only part of the KoDF dataset we were able to utilize was its adversarially-attacked subset.

\textbf{Models.} We use an AdaFace~\cite{adaface} model pre-trained on MS1MV2 \cite{arcface}, MS1MV3 \cite{lightweight} and WebFace4M \cite{WebFace260M} to extract per-frame identity features and use a transformer encoder to aggregate the identity in a temporal window for our visual features (both $\mu$ and $\gamma$). For the audio features ($\alpha$) we use a pre-trained Whisper~\cite{whisper} encoder trained on 680k hours of labeled audio-text data as our audio feature extractor. The entire model (including pretrained layers) are trained during our training with intra- and cross-modal losses. Details of our data preprocessing and hyperparamters are provided in Appendix.

\textbf{Metrics.} We use the standard Area Under the Curve (AUC) and Average Precision (AP) as performance metrics. These metrics are widely used since these two indicators do not impose threshold restrictions~\cite{self_supervised_video_forensics_by_audiovisual_anomaly_detection, Audio_visual_POI_deepfake_detection}. Furthermore, they effectively capture the model's performance in scenarios involving imbalanced datasets, accurately assessing the model's ability to detect true positives.

\begin{table*}[t]
  \caption{AUC ($\%$) and AP ($\%$) of our methods (Intra-modal, Cross-modal, and Intra-Cross-modal) on FakeAVCeleb\cite{fakeav}. Our combined method outperforms the unsupervised state-of-the-art, and reaches within $0.3$ absolute percentage points of best supervised method on average (AVG-FV).}

  \renewcommand{\arraystretch}{1.3}
  \centering
  \resizebox{\textwidth}{!}{

\begin{tabular}{cccccccccccccccccccccc}
\hline
\toprule[1.1pt]
\multirow{3}{*}{}             & \multirow{3}{*}{Method}  & \multirow{3}{*}{Modality} & \multirow{3}{*}{Pretrained Dataset} & \multicolumn{18}{c}{Catogory}                                                                                                                                                                                                               \\ \cline{5-22} 
                              &                          &                           &                                     & \multicolumn{2}{c}{RVFA}        &  &  & \multicolumn{2}{c}{FVRA-WL}     &  & \multicolumn{2}{c}{FVFA-FS}   &  & \multicolumn{2}{c}{FVFA-GAN}  &  & \multicolumn{2}{c}{FVFA-WL}     &                       & \multicolumn{2}{c}{AVG-FV}    \\ \cline{5-6} \cline{9-10} \cline{12-13} \cline{15-16} \cline{18-19} \cline{21-22} 
                              &                          &                           &                                     & AP             & AUC            &  &  & AP             & AUC            &  & AP            & AUC           &  & AP            & AUC           &  & AP             & AUC            &                       & AP             & AUC           \\ \cline{1-10} \cline{12-13} \cline{15-16} \cline{18-19} \cline{21-22} 
\multirow{5}{*}{\rotatebox[origin=c]{90}{Supervised}}   & Xception \cite{Xception}                & V                         & ImageNet                            & -              & -              &  &  & 88.2           & 88.3           &  & 92.3          & 93.5          &  & 67.6          & 68.5          &  & 91.0           & 91.0           & \multicolumn{1}{c|}{} & 84.8           & 85.3          \\
                              & LipForensics \cite{lipForenscis}            & V                         & LRW                                 & -              & -              &  &  & \textbf{97.8}  & \textbf{97.7}  &  & 99.9          & 99.9          &  & 61.5          & 68.1          &  & 98.6           & 98.7           & \multicolumn{1}{c|}{} & 89.4           & 91.1          \\
                              & AD DFD  \cite{joint_audio_visual_deepfake_detection}                 & AV                        & Kinetics                            &         74.9       &         73.3       &  &  & 97.0           & 97.4           &  & 99.6          & 99.7          &  & 58.4          & 55.4          &  & \textbf{100.}  & \textbf{100.}  & \multicolumn{1}{c|}{} & 88.8           & 88.1          \\
                              & FTCN   \cite{FTCN}                  & V                         & -                                   & -              & -              &  &  & 96.2           & 97.4           &  & \textbf{100.} & \textbf{100.} &  & 77.4          & 78.3          &  & 95.6           & 96.5           & \multicolumn{1}{c|}{} & 92.3           & 93.1          \\
                              & RealForensics   \cite{RealForensics}         & V                         & LRW                                 & -              & -              &  &  & 88.8           & 93.0           &  & 99.3          & 99.1          &  & \textbf{99.8} & \textbf{99.8} &  & 93.4           & 96.7           & \multicolumn{1}{c|}{} & \textbf{95.3}  & \textbf{97.1} \\ \midrule[0.7pt]
\multirow{7}{*}{\rotatebox[origin=c]{90}{Unsupervised}} & AVBYOL \cite{AVBYOL}                  & AV                        & LRW                                 & 50.0           & 50.0           &  &  & 73.4           & 61.3           &  & 88.7          & 80.8          &  & 60.2          & 33.8          &  & 73.2           & 61.0           & \multicolumn{1}{c|}{} & 73.9           & 59.2          \\
                              & VQ-GAN  \cite{VQ-GAN}               & V                         & LRS2                                & -              & -              &  &  & 50.3           & 49.3           &  & 57.5          & 53.0          &  & 49.6          & 48.0          &  & 62.4           & 56.9           & \multicolumn{1}{c|}{} & 55.0           & 51.8          \\
                              & A-V Anomaly  \cite{self_supervised_video_forensics_by_audiovisual_anomaly_detection}            & AV                        & LRS2                                & 62.4           & 71.6           &  &  & 93.6           & 93.7           &  & 95.3 & 95.8 &  & 94.1 & 94.3 &  & 93.8           & 94.1           & \multicolumn{1}{c|}{} & 94.2           & 94.5 \\  
                              & A-V Anomaly  \cite{self_supervised_video_forensics_by_audiovisual_anomaly_detection}            & AV                        & LRS3                                & 70.7           & 80.5           &  &  & 91.1           & 93.0           &  & 91.0 & 92.3 &  & 91.6 & 92.7 &  & 91.4           & 93.1           & \multicolumn{1}{c|}{} & 91.3           & 92.8 \\ \cline{2-22} 
                              & Ours (Intra-modal)                    & V                         & VoxCeleb2                                & -              & -              &  &  & 94.96           & 67.99           &  & 96.98          & 86.65          &  & 98.33          & 90.65          &  & 94.09           & 66.15           & \multicolumn{1}{c|}{} & 96.09           & 77.86         \\ 
                              & Ours (Cross-modal) & AV                        & VoxCeleb2                           & \textbf{99.68} & \textbf{99.65} &  &  & \textbf{99.37} & \textbf{95.98} &  & \textbf{98.74}         & 95.66        &  & \textbf{98.81}         & \textbf{94.58}         &  & \textbf{99.38} & \textbf{96.25} & \multicolumn{1}{c|}{} & \textbf{99.08} & \textbf{95.62} \\ 

                              & \textbf{Ours (Intra-Cross-modal)}                    & AV+V                         & VoxCeleb2                                & \textbf{98.49}             & \textbf{99.41}              &  &  & \textbf{99.34}           & \textbf{95.96}           &  & \textbf{99.27}          & \textbf{97.71}          &  & \textbf{99.43}         & \textbf{97.59}          &  & \textbf{99.29}           & \textbf{95.99}           & \multicolumn{1}{c|}{} & \textbf{99.33}           & \textbf{96.81}         \\

                            \bottomrule[1.1pt]

\end{tabular}
}

    \label{tab:results}
\end{table*}

\subsection{Main Results}
\label{sec:main_results}
Following the evaluation setup of  \cite{self_supervised_video_forensics_by_audiovisual_anomaly_detection}, we separately consider the five categories of FakeAVCeleb dataset (RVFA, FVRA-WL, FVFA-WL, FVFA-FS, FVFA-GAN). Each category includes deepfake videos generated using the same method. The evaluation process allows us to gain clear insights into how our method performs under different types of deepfake generations, which can also substantiate our conjecture that \textit{deepfake generators will inevitablly leave a detectable trace}.

The result of evaluation on FakeAVCeleb is shown in \cref{tab:results}. We observe that our overall consistency method (Intra-Cross-modal) which combines the strengths of both intra- and cross-modal scores, consistently achieves a significant improvement over previous state-of-the-art unsupervised methods across all categories, and is only $0.3\%$ lower than the best-performing \emph{supervised} method on average (AVG-FV). Note that our method achieves this performance \emph{without seeing any of the identities in the FakeAVCeleb dataset, or any of the various deepfake generations used in this dataset}, during its training. Our method is also does not require/have access to any pristine (real) videos during inference.

An key observation is that while our combined method (Intra-Cross-modal) achieves state-of-the-art performance, the performance of the two sub-scores (Intra-modal and Cross-modal) varies depending on the deepfake type and show complementary strengths (the worst AUC of the Cross-modal method is on FVFA-GAN, which is where the Intra-modal method has its best AUC). This observation validates  our theoretical motivation that there is a trade-off between identity consistency and motion consistency, and therefore detecting both types of inconsistencies should be considered. We discuss the separate performance of the Intra-modal and Cross-modal methods in~\cref{sec:intra_results}.

\begin{figure*}[t]
    \centering
    \includegraphics[width=1\textwidth]{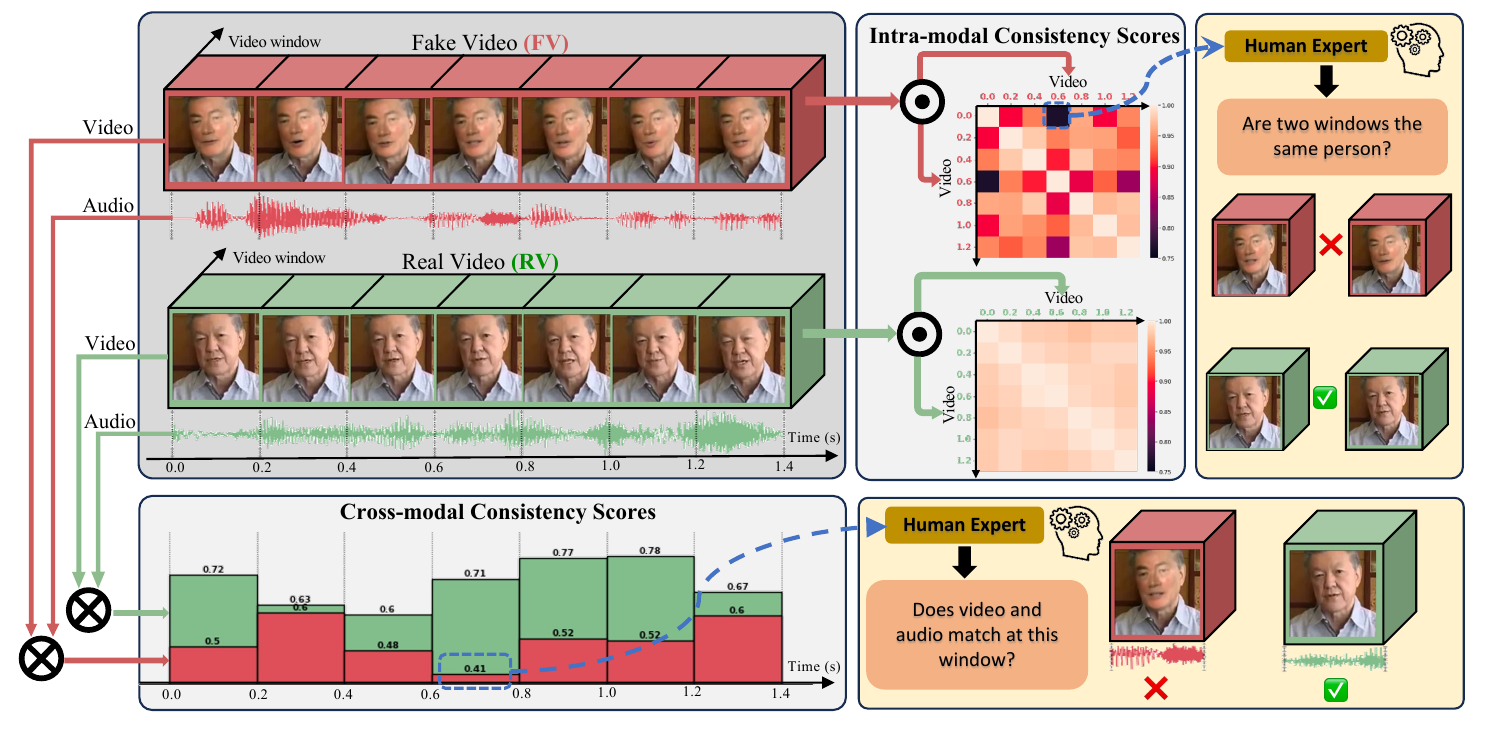}
    \caption{The explainability of the proposed methods using intra-modal consistency loss and cross-modal consistency loss for two samples in FakeAVCeleb. When the method decides that a given video is fake due to its average score being lower than a threshold \emph{(light-gray boxes)}, it can provide the portions of the video with the minimum consistency score to a human expert as explanation \emph{(yellow boxes)}, and the expert can verify the method's decision through manual comparison.}
    \label{fig:inter-cross-scores}
    \vspace{-.3in}
\end{figure*}

Furthermore, as illustrated in~\cref{fig:inter-cross-scores}, the predictions of our method are explainable by design. For example,  the video segments with highest mismatch score can be optionally provided to a human expert as explanations of ``fakeness'', and are verifiable through further analysis by such forensic experts.

\subsection{Generalization to Audio/Video Compression, and KoDF Adversarial Attacked Dataset}
\label{sec:robustness}

\begin{table*}[t]
    \caption{AUC ($\%$) comparison of our Intra-Cross-modal method on high/low quality compressed FakeAVCeleb dataset, and attacked KoDF dataset. The best and second-best results of unsupervised methods are in bold and underlined, respectively.}
  \renewcommand{\arraystretch}{1.2}
  \centering
  {

\setlength{\tabcolsep}{8pt}
\begin{tabular}{cccccc}
\hline
 &\multirow{2}{*}{Methods}& \multicolumn{2}{c}{FakeAVCeleb} & \multicolumn{1}{c}{KoDF} & \multirow{2}{*}{Training Dataset}\\ 
                                     &      & HQ   & LQ  & Attacked &  \\ \hline
{\multirow{5}{*}{\rotatebox[origin=c]{90}{Supervised}}} & Seferbekov\cite{seferbekov}     & 98.6 & 61.7 & 61.5&     DFDC     \\
                & FTCN\cite{FTCN}           & 84.0 & 37.6 & 58.3 &    -    \\
             & LipForensics \cite{lipForenscis}  & 97.6 & 58.3&54.8 &    LRW        \\
            & Real Forensics\cite{RealForensics} & 88.3 & 52.9 &  55.5&  LRW        \\
                & MDS-based FD\cite{MDS-basedFD}   & 64.7 & 61.1 & 55.4 &  LRW     \\ 
 & Joint AV\cite{joint_audio_visual_deepfake_detection}       & 55.1 & 55.2 & 47.1 &   DFDC    \\ \hline
{\multirow{5}{*}{\rotatebox[origin=c]{90}{Unsupervised}}}                  & ICT \cite{ICT}           & 68.2 & 66.9 & 61.5 &    MS-Celeb-1M       \\
               & ICT-Ref\cite{ICT}      & 71.9 & 71.2 & 78.0 &  MS-Celeb-1M      \\
                & ID-Reveal \cite{idreveal}      & 70.2 & 70.8 & 73.4 &    VoxCeleb2    \\
                & POI-Forensics \cite{Audio_visual_POI_deepfake_detection}  & \underline{94.1} & \textbf{94.4} &  \underline{80.5} &   VoxCeleb2     \\ \cline{2-6} 
                & Ours  (Intra-Cross-modal)        & \textbf{95.5} & \underline{85.9}  & \textbf{84.2}{\tiny ±1.1}  &    VoxCeleb2   \\ \hline
\end{tabular}
}

    \label{tab:hq_lq_table}
\end{table*}

In this section, we study the generalization capability of our proposed method (Intra-Cross-modal) to varying video and audio qualities, as well as to other spoken languages and adversarial attacks. We follow the evaluation settings of \cite{Audio_visual_POI_deepfake_detection}. We could not include AV-Anomaly \cite{self_supervised_video_forensics_by_audiovisual_anomaly_detection} in these evaluations due to lack of public inference and training code.

\textbf{Compression Study.} We construct a high quality and low quality version of the FakeAVCeleb dataset. In the High Quality (HQ) setting, the video is compressed using H.264 encoding with factor 23, and the audio is the same as original (44.1KHz). In the Low Quality (LQ) setting, the video is compressed using H.264 encoding with factor 40 and the audio is sampled with a sample rate of 16KHz. In ~\cref{tab:hq_lq_table}, we observe that our method outperforms all other unsupervised methods in the HQ compression setting (small video compression, no audio compression). However, in the LQ setting, our method is second to POI-Forensics, while still outperforming all other methods. However, it is important to note that POI-Forensics \emph{requires} access to pristine videos during testing, which it can compress to the same LQ setting and directly compare against.

\textbf{KoDF Study.} We also evaluated our method using the adversarially attacked subset of the KoDF\cite{KoDF} (both real and deepfake videos are adversarially attacked using the Fast Gradient Sign Method (FGSM)\cite{fgsm_adversarial_attack}). In this setting, POI-Forensics~\cite{Audio_visual_POI_deepfake_detection} randomly sampled 276 original videos and 544 synthesized videos as positive and negative samples for evaluation. The specific method for random selection of data was not disclosed by \cite{Audio_visual_POI_deepfake_detection}. Therefore, for a robust and fair comparison, we repeat our evaluation 1000 times following their setting, each time with new random choice of 276 original and 544 synthesized videos from KoDF, which contains 5365 original and 15044 synthesized videos, and report the average and one standard deviation of our method's AUC (\%). As shown in ~\cref{tab:hq_lq_table}, we observe that our method outperforms all supervised and unsupervised methods on this challenging task, where not only the language of the videos are completely different than training, but all the samples are also adversarially attacked. This empirical evidence serves as an additional validation of the theoretical motivation, that there is an inevitable consistency (in either identity or motion) of deepfake videos, as discussed in \cref{sec:theory}.

\subsection{Why Does Intra-Modal Performs Worse than Cross-Modal?}
\label{sec:intra_results}

\begin{figure*}[t]
    \centering
    \includegraphics[trim={5 5 5 5},clip, width=0.8\textwidth]{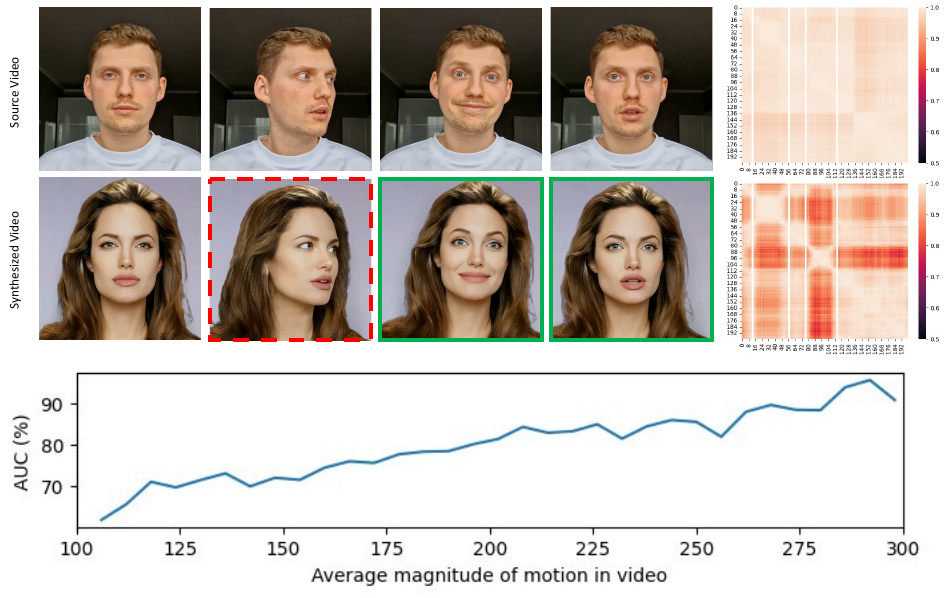}
    \caption{The similarity matrices ($M_{intra}$) of the Intra-modal method clearly show the stronger temporal fluctuations of identity in deepfake \emph{(middle)} compared to real \emph{(top)}.
    Intra-modal method AUC on FaceAVCeleb sorted based on the magnitude of motion in videos shows increasing performance with increasing motion in videos \emph{(bottom)}.}
    \label{fig:auc_motion_distance}
    \vspace{-.25in}
\end{figure*}

We observe in \cref{tab:results} that our Intra-modal method performs worse than our Cross-modal method on FakeAVCeleb. The reason can be understood from our theoretical analysis. A deepfake generation method must produce identity inconsistencies \emph{if it transfers the motion exactly}. In other words, if the deepfake generation method does not have to transfer motion (imagine training a deepfake method on a video dataset that has no motion at all, and only still portraits), then the generation method can transfer the identity perfectly, and consequently our Intra-modal identity consistency method will be completely ineffective. As such, whether Intra-modal method or Cross-modal method perform better, depends on how well the deepfake generation method is transferring motion.

If the data contains little facial motion (such as FakeAVCeleb, where most videos are stationary, calm, frontal-shot interviews) a deepfake generation method will incur little penalty for not perfectly transferring motion, and consequently the trained deepfake method will be at the motion-inconsistency side of the trade-off predicted by our theory. Therefore, the Intra-modal method will perform worse that the Cross-modal method---which is detecting motion inconsistency. However, note that it is precisely for this reason that we proposed both methods, because they are complementary in their capabilities, and which one performs better alone depends on the particular deepfake generation method.

To support the above argument regarding the behavior of our Intra-Modal method, we provide both quantitative and qualitative evidence  in ~\cref{fig:auc_motion_distance}. To illustrate the other side of the theoretical trade-off, we show that our Intra-modal method can clearly discover mismatched identity in videos of a deepfake model that is trained to strongly regress the source motion ~\cite{drobyshev2022megaportraits} (thereby losing identity consistency). Furthermore, to illustrate how the Intra-modal method is very effective in particular deepfakes that require substantial motion matching (\ie, generating a mismatched motion would incur a great loss for the deepfake generation method), in~\cref{fig:auc_motion_distance} we show that the AUC of the Intra-modal method increases with the average magnitude of motion in videos of FakeAVCeleb (measured for each video as the $L_2$ norm of the motion vector).

\section{Conclusion}
\label{sec:conclusions}
In this work, we introduced a novel theoretical analysis that suggests motion or identity inconsistencies are inevitable in deepfake videos that transfer the motion of a source video to a target video. We provided mathematical and empirical evidence for our theoretical claim, and then based on this theory, we proposed unsupervised methods to explicitly detect and pinpoint identity and motion inconsistencies in deepfakes, namely the Intra-modal method and Cross-modal method, and combined them into the Intra-Cross-modal method that achieved a new state-of-the-art performance on the FakeAVCeleb dataset. We also showed that our method generalizes to videos of other languages (KoDF Korean speech deepfake dataset), adversarial attacks, and small video compression. Furthermore, compared to existing deepfake detection methods, our method is more scalable because it does not require any pristine (real) videos during inference, generalizable because it only trains on real videos, reliable because it does not explicitly use likelihood estimation in high dimensions, and explainable because it explicitly discovers verifiable inconsistent segments in a video. Our findings also reveal several interesting directions for future research. First, directly measuring the terms in the proposed information-theoretical upper bound in~\cref{sec:theory} for various deepfake methods can empirically verify the bound and reveal interesting trends in deepfake videos. Second, we expect the proposed Intra-modal method could be further improved by building inductive biases into the architecture that encourages attending to fine visual details. Finally, while we currently only consider talking-head videos, the proposed consistency losses could be applied to other video types such as to full-body videos.

\clearpage  %

\bibliographystyle{splncs04}
\bibliography{main}
\end{document}